# OVERVIEW OF SATELLITE IMAGE RECOGNITION MODELS[1]

## 2022 A.N. Averkin a  S.A. Yarushev[a]

[a]*Plekhanov Russian University of Economics, Moscow, Russia*

*e-mail:averkin2003@inbox.ru, sergey.yarushev@icloud.com*

In this article, the analysis of existing models of satellite image recognition was carried out, the problems in the field of satellite image recognition as a source of information were considered and analyzed, deep learning methods were compared, and existing image recognition methods were analyzed. The results obtained will be used as a basis for the prospective development of a fire recognition model based on satellite images and the use of recognition results as input data for a cognitive model of forecasting the macro-economic situation based on fuzzy cognitive maps.

**Introduction.** Satellite imagery is becoming an increasingly serious tool for visualizing and mapping the entire planet. Collecting images is getting faster every day, and access to images is becoming more convenient. [1] Based on all this, the quality of satellite images is also improving, and in the public domain there are gradually more and more images that can be used for analytical purposes. For example, to train a neural network to recognize objects on satellite images, which in turn can also be used to solve all sorts of applied problems. One of these tasks is the classification and analysis of natural disasters on satellite images. Given the trends in the development of satellites and satellite imagery, this method of forecasting is more relevant than ever.

The large-scale development of artificial intelligence (AI) systems, including applications based on artificial neural networks, opens up the widest possibilities for their use in various fields, from emotion recognition systems to predictive analytics systems, applications in medicine and the military field.  But, at the same time, existing systems and applications have one significant drawback in common - the inability to interpret the results obtained and the decisions made. The well-known problem of the so-called black box imposes significant restrictions on the use of such systems, including legislative ones, since it is impossible to trace the course of decision-making by a neural network.  Althoughsatellite imagery has become a truly important and convenient source of information, this source still has one major drawback. This drawback is that finding enough satellite imagery in open sources for training sampling is a big problem [2] ]. Of course, if government agencies or science are interested in this issue, then this problem becomes completely irrelevant, because in this case you can get access to the huge databases of satellite images that most states with a developed space industry possess.

[1] This research was performed in the framework of the state task in the field of scientific activity of the Ministry of Science and Higher Education of the Russian Federation, project "Development of the methodology and a software platform for the construction of digital twins, intellectual analysis and forecast of complex economic systems", grant no. FSSW-2020-0008.

**1. The problem of data collection for the classification of satellite images**. One of the most important steps in the entire process of implementing a classification model is data collection. It is on this stage that the continued success of the whole work depends, because no matter how powerful the deep learning algorithms themselves are, they are not able to provide a qualitative result with a poor initial data set. It is errors and problems with the initial data for analysis that most often cause failures in practice.

As for the collection of data in the context of the classification of satellite images, this source of information is gaining more and more popularity every year. And not the last role in this is played by the peculiarity and a kind of uniqueness of satellite images as a source of data for forecasting. The obvious advantages of satellite imagery include:

• satellite images are a source of unique information and can be used for forecasting in areas in which no other source is suitable for such forecasting (for example, recognition of natural disasters; recognition of forest park plantations, forecasting the melting of glaciers, etc.);

• cheapness of the source of information (despite the fact that the common man has to be content with satellite images that are in the public domain, there are government agencies that have full access to all satellite images of satellites of this state);

However, it is necessary to mention the main drawback of this data source - we can highlight the fact that in open sources it is not so easy to find the number of images necessary for training sampling. Especially when it comes to already marked data. During the markup process, special metadata is added to the original image or video - these are certain tags that carry information about some specific properties of an object [3].

Training a neural network on satellite images allows you to give the network the ability to classify natural disasters: droughts, fires, floods. This information will allow interested persons to take timely measures to eliminate the consequences of natural disasters or even prevent them.

**2. Analysis of existing satellite image recognition models**. Let's look at existing solutions in the field of satellite image recognition. One of these solutions is the system of recognizing houses on satellite images from the company "Yandex". Satellite images are one of the main data sources for filling the service "Yandex.Maps". At the heart of this solution is a convolutional neural network that solves the problem of semantic segmentation, that is, the network determines whether each point in the satellite image belongs to the house. At the output, a set of specific rectangles is obtained: two sides of which are vertical, and two are horizontal. Houses are usually rotated relative to the axes of the image, and some buildings also have a complex shape. Having received a mask from a satellite image, large clusters of points that belong to houses stand out. Next, they are assembled into connected regions, and the boundaries of the regions are represented in vector form in the form of polygons. Obviously, the mask won't be absolutely accurate, which means that homes that stand close together can stick together in one cohesive area. In order to solve this problem, the network was additionally trained. Thus, the network learned to find the edges (borders of houses) in the image and separate the buildings that are glued together. A visual diagram of the work is presented in Figure 1.

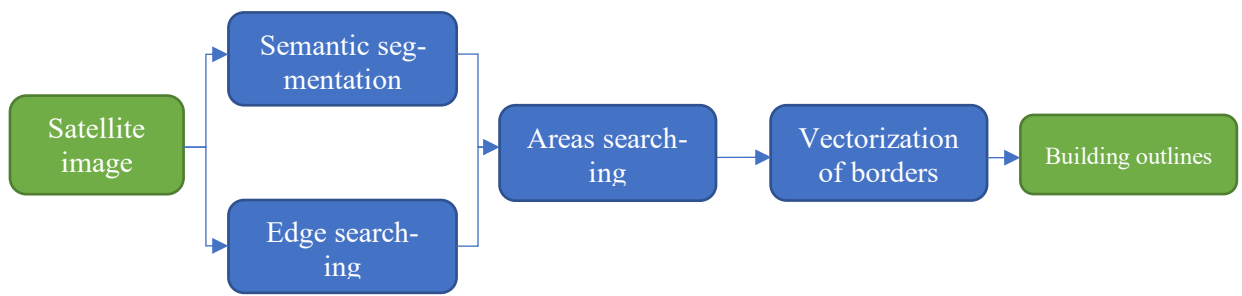

Figure 1 – Scheme of the neural network in the solution from the company "Yandex"

As a dataset, Yandex uses its own data with already pre-marked objects, which greatly simplifies the task. However, there is still one problem. The fact is that if a person begins to look for houses on a satellite image, then the first thing that comes across his eyes is the roofs of houses. But it is worth noting that all houses have different heights, and the satellite can shoot the same location, but from completely different angles - and if you place a polygon corresponding to the roof of the house on the vector map, then nothing guarantees us that when updating the picture, the roof of the house will not shift. At the same time, the foundation of the buildings is dug into the ground and its location in the picture does not depend on the angle from which the shooting is carried out. Therefore, the houses on the vector "Yandex.Map" are marked exactly "according to the foundations". This is correct, but for the task of segmenting images, it is better to teach the network to look for roofs, because the hope that the network will learn to recognize the foundations is very small, so in the dataset everything should be marked on the roofs. It turns out that in order to create a good dataset, it was necessary to learn how to shift the vector marking of houses from foundations to roofs. To do this, Yandex used 2 approaches: "raster" and "geometric".

Let's take a closer look at the "raster" approach. First, the vector map is rasterized (polygons of houses are drawn with white lines on a black background), then using the Sobel filter (a discrete operator that calculates the approximate value of the gradient of the brightness of the image), the edges on the satellite image are highlighted. And then there is the displacement of the two images depending on their location from each other, which maximizes the correlation between them. Ribs after applying the Sobel filter have a high level of noise, therefore, you should not apply this approach to one single house, because, most likely, an unsatisfactory result will be obtained. However, this method manifests itself well in areas with buildings of the same height: if you look for displacement at once over a large area of the image, the result will be more stable [4]. The result of applying the raster approach is shown in Figure 2.

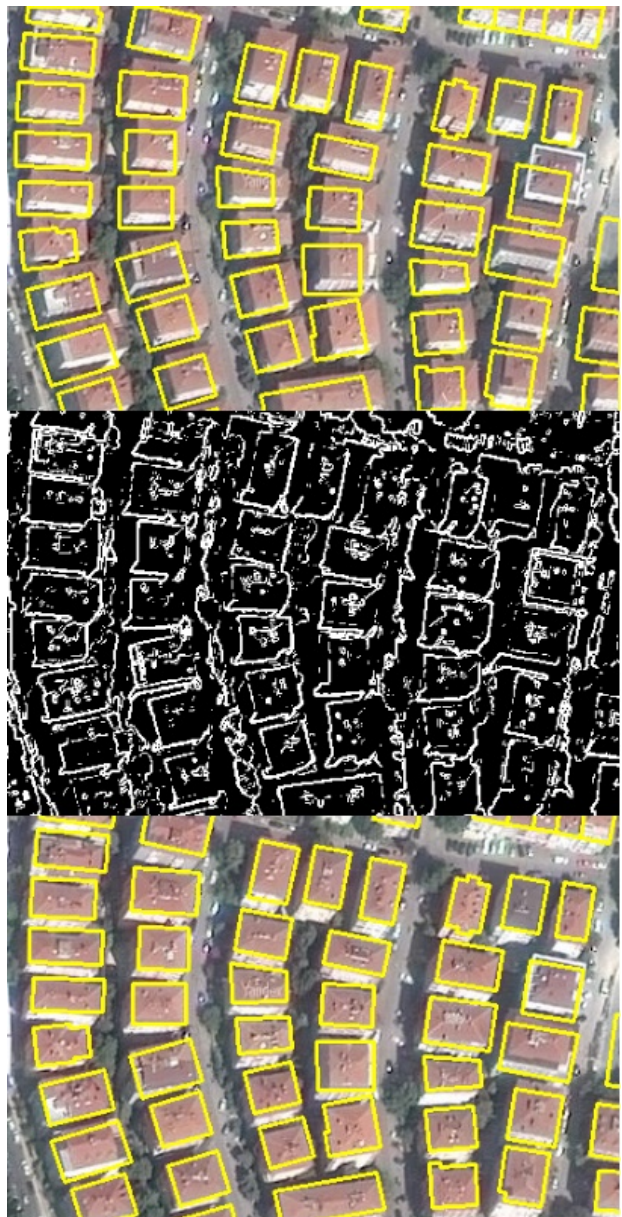

Figure 2 – The result of the raster approach

Now let's look at the "geometric" approach. If the territory is built up not with the same type, but with a variety of houses, then the previous method will not work. But sometimes on the vector map of Yandex, the height of the building and the position of the satellite at the time of shooting are already known. Thus, with the help of geometric calculations, it becomes possible to calculate where and by what distance the roof will move relative to the foundation. This method allowed to improve the dataset in areas with high-rise buildings. The meaning of the geometric approach is shown in Figure 3.

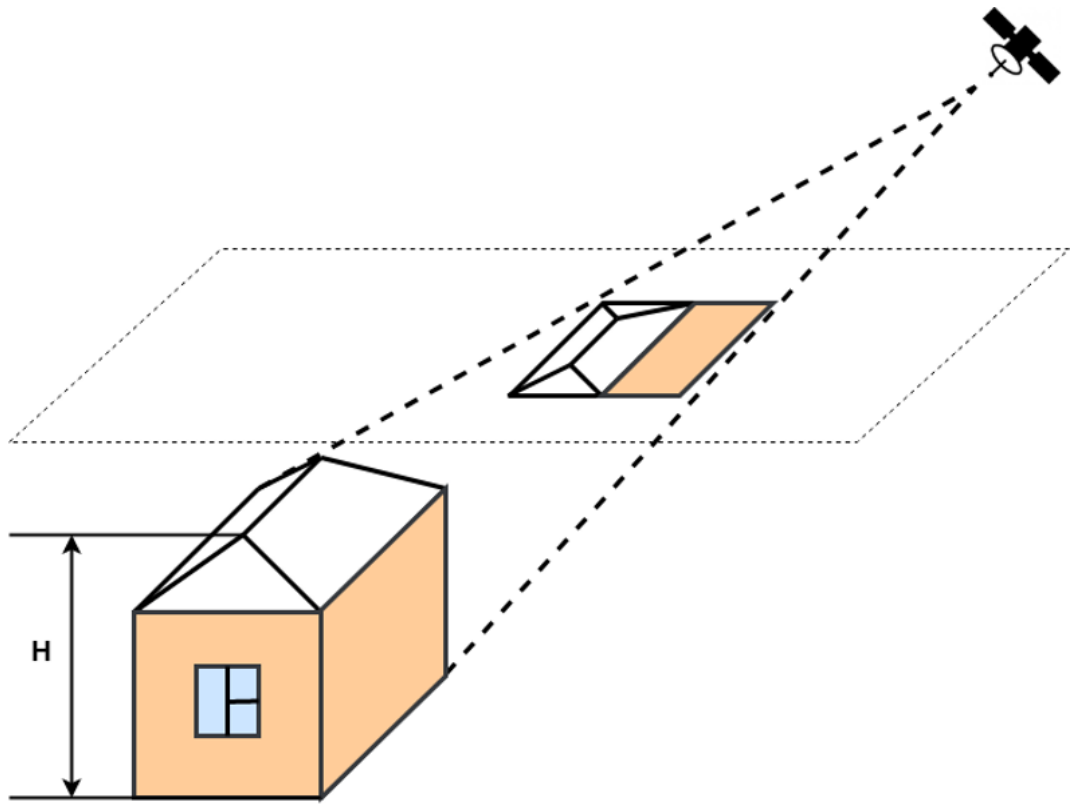

Figure 3 – "Geometric" approach.

The result of applying the geometric approach is presented in Figure 4.

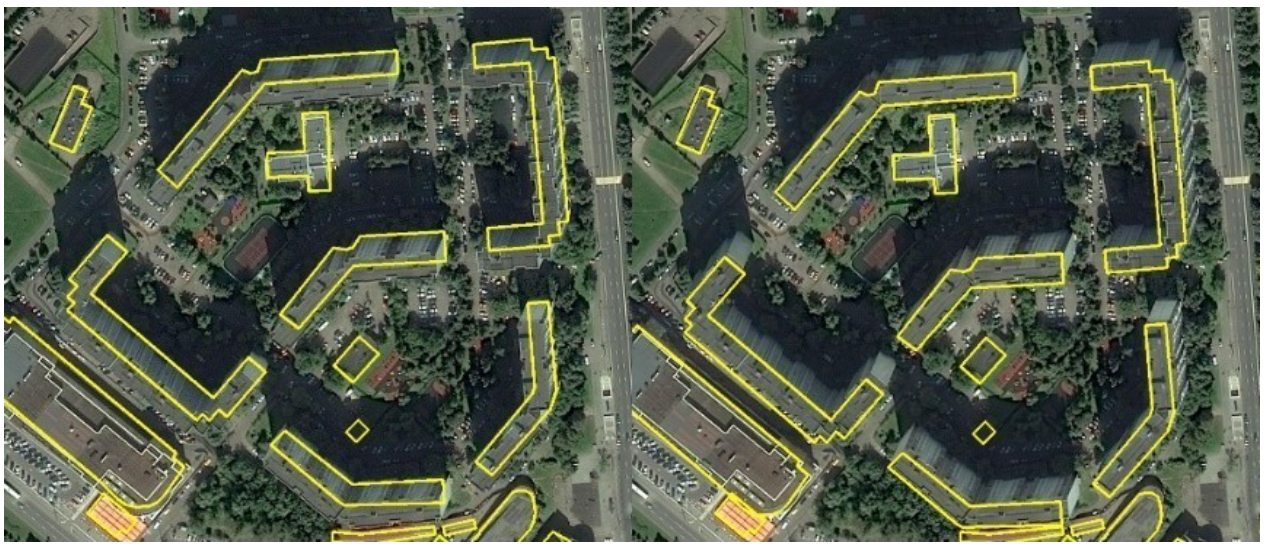

Figure 4 - The result of the application of the geometric approach.

As a result, a sufficient number of satellite images were obtained, well marked on the roofs. So,

there was a chance to train a neural network, which was done. The input data of the convolutional neural network was a satellite image and shifted masterized markup. The output was a two-dimensional vector: vertical and horizontal displacements. The input and output data of the convolutional neural network used in Yandex to form Yandex.Maps are presented in Figure 5.

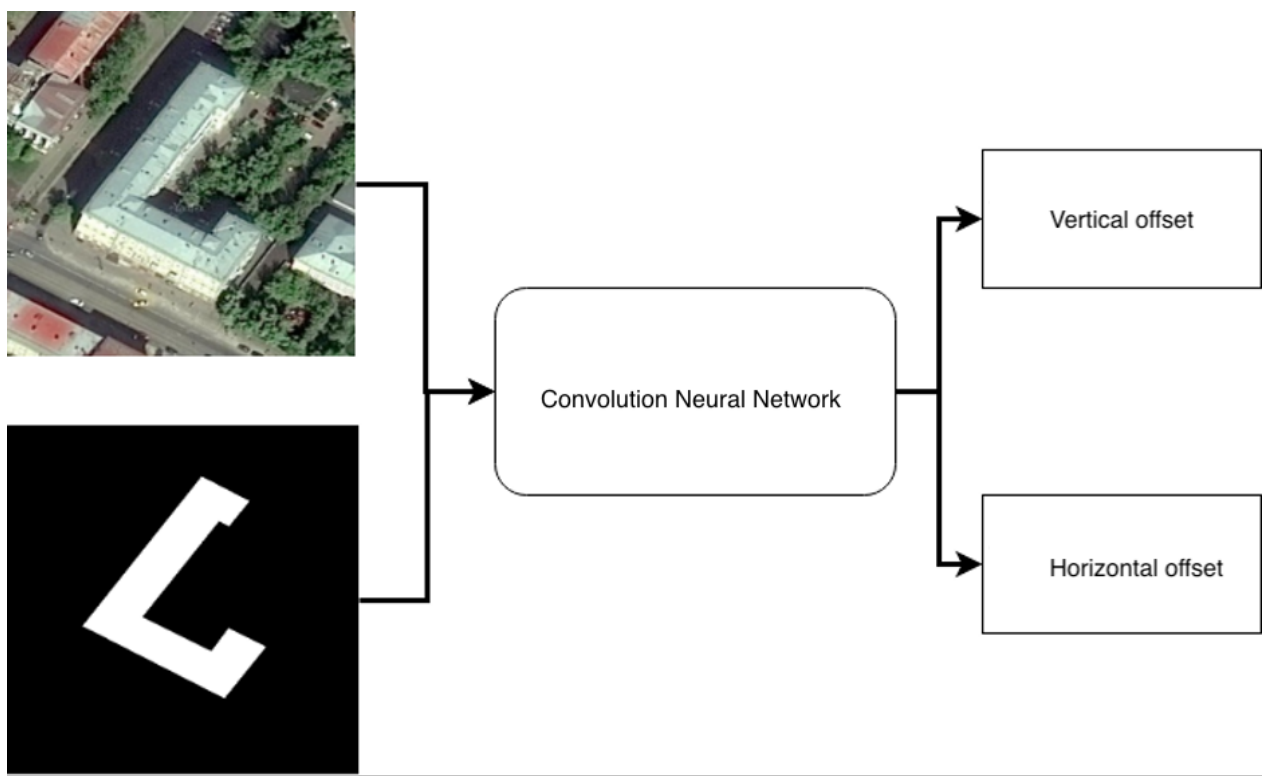

Figure 5 – Convolutional Network Inputs and Outputs

The network architecture is a U-Net-like architecture with a generalized Intersection Over Union feature as an error function.

**Conclusion.** This article presents research in the field of fire detection on satellite images and considers the prospects for using the resulting recognition model to solve an important, socially significant task of predicting the consequences of natural disasters on the macroeconomics of the affected region. The paper presented the existing ready-made satellite image recognition systems, as well as possible architectures based on deep neural networks to solve this problem.